\pdfoutput=1
\documentclass[11pt]{article}
\usepackage[utf8]{inputenc}
\usepackage[T1]{fontenc}
\usepackage{newtxtext,newtxmath}
\usepackage{amsmath}

\usepackage{amssymb}
\usepackage{graphicx,booktabs,tabularx,array,microtype}
\usepackage{flafter,caption}

\usepackage[margin=1in]{geometry}
\usepackage[table]{xcolor}
\usepackage{tikz,titlesec,etoolbox}
\usetikzlibrary{arrows.meta,positioning,calc,fit,backgrounds}
\usepackage[authoryear,round]{natbib}
\usepackage{hyperref}
\usepackage{cleveref}
\hypersetup{colorlinks=true,linkcolor=blue!50!black,citecolor=blue!50!black,urlcolor=blue!50!black,
 pdftitle={SymbolicLight V2: Hybrid Neuromorphic Architecture and Sparse Execution for Low-Energy Language Inference},pdfauthor={Ting Liu},pdfkeywords={hybrid neuromorphic language architecture; event-driven computation; FPGA; sparse integer execution; energy per generated token}}

\definecolor{slblue}{HTML}{3568A8}
\definecolor{slgreen}{HTML}{168579}
\definecolor{slpale}{HTML}{EDF4F6}
\titleformat{\section}{\large\bfseries\color{black}}{\thesection}{0.7em}{}
\titleformat{\subsection}{\normalsize\bfseries}{\thesubsection}{0.7em}{}

\arrayrulecolor{black}

\newcommand{\snew}{SL\mbox{-}V2}
\newcommand{\czero}{SL\mbox{-}V2\mbox{-}Cont}
\newcommand{\srepair}{SL\mbox{-}V2\mbox{-}Softmax}
\newcommand{\gnew}{GPT\mbox{-}2~194M}
\newcommand{\tps}{tok/s}
\newcommand{\impldense}{resident\mbox{-}dense}
\newcommand{\implgather}{sparse\mbox{-}gather}
\newcommand{\implkv}{partial\mbox{-}KV}
\newcolumntype{Y}{>{\raggedright\arraybackslash}X}
\title{\textbf{SymbolicLight V2: Hybrid Neuromorphic Architecture and Sparse Execution for Low-Energy Language Inference}}
\author{Ting Liu\\SymbolicLight Research\\Foshan, Guangdong, China\\\texttt{research@symboliclight.com}}
\date{September 7, 2026}
\begin{document}
\maketitle
\begin{abstract}
Event-driven language inference saves energy when the execution system turns zero activations into omitted computation and memory access. We present SymbolicLight V2, a hybrid neuromorphic language architecture combining sparse event computation with continuous-state processing. V2 extends V1's spike-gated dual-path design with graded signed events at additional projections and softmax-free local attention. An Alveo U50C field-programmable gate array (FPGA) implements an end-to-end digital fixed-point inference prototype of the 194M-parameter model, alongside a sparse integer ARM implementation. Across three same-checkpoint FPGA implementations at 175\,MHz, active-row weight gathering and valid-state key/value (KV) cache loading raise continuous decode throughput for a 32-token prefix and 128 generated tokens from 474.6 to 643.2 tokens/s, while reducing estimated gross card energy from 0.06087 to 0.04407\,J per generated token (27.6\%). Complete-request energy, including prefill, falls by 24.4--27.7\% across three tested prefix lengths. An independent idle-split measurement attributes 82.8\% of gross card energy to the loaded-idle share, explaining the energy value of shorter token latency. Against the recorded RTX~5090 compiled 32-bit floating-point (FP32) baseline, the integer FPGA deployment uses 89.1\% less estimated card energy during short-context decode; the comparison uses different arithmetic precisions and does not use the lowest-energy tested GPU configuration. On four Cortex-A76 cores of a ROCK~5T, complete requests reach 65.4 generated tokens/s at 9.80\,W and 0.151\,J/token at the adapter's AC input. The results show how sparse execution of a hybrid neuromorphic language architecture lowers measured inference cost through event-aware computation and data movement. These mechanisms also provide a design basis for other dedicated V2 implementations: increasing throughput by a larger factor than the change in active power reduces energy per generated token. Evaluation fixes the deployed checkpoint, whose quality trails a same-budget dense control; it does not establish equal-quality efficiency.
\end{abstract}

\section{Introduction}
\label{sec:intro}
Activation sparsity reduces inference energy when the execution system avoids the corresponding arithmetic and memory traffic. SymbolicLight V1 combined binary Leaky Integrate-and-Fire (LIF) dynamics, a continuous residual stream, and a dual-path temporal mixer \citep{liu2026symboliclightv1}. A recurrent decay path maintained temporal state while local attention accessed recent context. V1 explicitly adopted a hybrid design, using the combination of discrete spikes and continuous internal dynamics as biological motivation. Although this model trained with sparse activations, its dense GPU implementation paid for work that those zeros could have removed.

V2 develops this lineage into a hybrid neuromorphic language architecture with executable sparsity. Here, hybrid neuromorphic denotes the cooperation of sparse event computation and continuous-valued state processing: events select projection contributions, while recurrent state and a continuous residual carry temporal and feature information. The prototype implements both using digital fixed-point arithmetic. V2 extends event coding to feed-forward up-projection and query/key/value (Q/K/V) inputs, eventizes the projected vectors again, and uses ReLU--$L_1$ local attention with linear position bias. Its deployment turns nonzero events into selected weight-row accesses: ARM loops skip zero-input contributions, and the FPGA gathers only the required rows from resident high-bandwidth memory (HBM).

The principal experiment holds the trained checkpoint fixed while improving FPGA execution. Replacing dense weight enumeration with active-row gathering, then loading only valid key/value (KV) cache state, increases p32/n128 continuous decode throughput by 35.5\% and reduces estimated gross card energy by 27.6\%. Here p32/n128 denotes a 32-token prefix followed by 128 generated tokens. Including prompt ingestion, energy falls by 24.4--27.7\% across the three tested prefix lengths. These are measured changes across successive implementations; the experiment does not isolate every hardware change or thermal effect.

The results also explain why latency matters to energy. Sparse projections expose KV-loading waits that were previously overlapped with computation. Reducing unnecessary state transfer recovers part of the execution benefit. Here, loaded idle means that model weights remain in device memory while no inference is running. In an independent measurement, loaded-idle power accounts for 82.8\% of the FPGA's gross energy per token. Faster generation therefore reduces the platform energy allocated to each output even when incremental energy changes little.

Two deployment comparisons place these results in context. Against the recorded RTX~5090 compiled-FP32 baseline, short-context integer FPGA decoding uses 89.1\% less estimated card energy, with precision and sensor boundaries stated alongside the comparison. On four Cortex-A76 cores, the same integer model completes short requests at 65.4 generated tokens/s and 0.151\,J/token at the board adapter's AC input. A comparison of six CPU and neural processing unit (NPU) deployments on the same board shows how this edge deployment trades throughput and energy as input length grows.

The paper makes three contributions:
\begin{enumerate}
\item \textbf{Hybrid neuromorphic architecture and sparse execution.} V1-derived dual paths combining extended graded-event projections with continuous-state processing, mapped to actual nonzero-input integer computation on ARM and active-row weight gathering on FPGA (\Cref{sec:arch,sec:hardware}).
\item \textbf{Same-checkpoint energy reduction.} A measured FPGA implementation progression that connects selective weight access and valid-state KV loading to higher throughput and lower gross energy, with an independent idle-share decomposition (\Cref{sec:results}).
\item \textbf{Cross-platform deployment evidence.} Complete-request ARM AC measurements, FPGA/GPU card-sensor comparisons, and explicit workload and measurement boundaries, supported by integer-reference validation (\Cref{sec:protocol,sec:results}).
\end{enumerate}
The evaluation focuses on execution of a fixed checkpoint. Model-quality diagnostics are summarized in \Cref{sec:scope} and reported in \Cref{app:quality}.

\section{From V1 to the V2 Hybrid Neuromorphic Architecture}
\label{sec:arch}
\subsection{The V1 starting point}
V1's two paths operate within each block: a first-order decay state and local attention, combined with a continuous residual stream and a feed-forward sublayer. Binary spike inputs feed the decay projection and the feed-forward down-projection; the LIF input encoder and the stateless threshold operations within later blocks have distinct roles. In contrast, Q/K/V and the feed-forward up-projection consume continuous inputs. Zero activations at the original spike sites therefore leave substantial dense computation elsewhere in the graph.

Per-element activity also cannot serve as a token-level attention mask. V1 measured an almost always active position mask: reducing the number of active elements did not remove whole keys. Its released implementation computed dense attention scores before applying the local mask, and cached decoding retained all historical K/V entries. The visible local window therefore did not bound the actual computation, nor did element-level sparsity reduce attention work proportionally. V2 consequently acts at projection inputs and within the attention arithmetic, instead of treating a token as inactive only when all its elements are zero.

\subsection{Graded signed events at projection inputs}
Let $x$ be the activation presented to an event encoder with positive threshold $\theta$ and positive integer magnitude bound $L$. The hard forward event is
\begin{equation}
 e_\theta(x)=\mathbf{1}[|x|\geq\theta]\,
 \operatorname{clip}\!\left(\operatorname{round}(x/\theta),-L,L\right).
 \label{eq:event}
\end{equation}
Here $\operatorname{round}$ rounds to the nearest integer with ties to even, $\operatorname{clip}$ bounds the result to $[-L,L]$, and $\mathbf{1}$ is the indicator function. A zero value denotes no event; a nonzero integer carries both sign and magnitude. Training uses a straight-through estimator with surrogate input derivative $\widehat{\partial e_\theta/\partial x}=1/\theta$, including the silent and saturated regions of the hard forward function. This follows the broader use of surrogate derivatives in spiking networks \citep{neftci2019surrogate}. V2 adds these encoders at the feed-forward up-projection input and at Q/K/V projection inputs in every block. Its selected configuration leaves the attention output-projection input continuous. The binary-spike-consuming decay and feed-forward down-projections inherited from V1 remain.

The mathematical event alphabet and the deployment payload width are distinct. The training configuration allows $L=255$, whose signed range is wider than signed INT8. The hardware representation uses signed 8-bit payloads with its specified quantization and saturation semantics. Accordingly, the hardware correctness claim is made against the integer deployment reference, not by assuming the entire training event alphabet fits into INT8.

Using column vectors throughout, let $e\in\mathbb R^{d_{\rm in}}$, $y\in\mathbb R^{d_{\rm out}}$, and $W\in\mathbb R^{d_{\rm out}\times d_{\rm in}}$. An event-consuming projection is
\begin{equation}
 y = We
   = \sum_{j\in\mathcal A(e)} e_j W_{:,j},
 \qquad \mathcal A(e)=\{j:e_j\ne0\}.
 \label{eq:gather}
\end{equation}
The active set $\mathcal A(e)$ selects columns of the mathematical matrix $W$. Hardware stores $W^{\mathsf T}$ by input index, so these columns correspond to stored weight rows; a zero event requires no row fetch. Since nonzero events have graded amplitudes, their execution generally still requires integer multiply-accumulate (MAC) operations. The savings come from omitted contributions and weight movement, not from treating every event as an addition-only binary spike.

\subsection{Event attention and the retained dual path}
The decay path preserves recurrent temporal state. For layer $\ell$ and binary input $s_{\ell,t}$, the vector $\alpha_\ell$ contains one learned sigmoid-parameterized decay coefficient per head, broadcast across that head's channels. The operator $\odot$ denotes elementwise multiplication, and a new session starts from zero state. Its recurrence is
\begin{equation}
 z_{\ell,t}=W_Ts_{\ell,t},\qquad
 h_{\ell,t}=\alpha_\ell\odot h_{\ell,t-1}+(1-\alpha_\ell)\odot z_{\ell,t}.
 \label{eq:decay}
\end{equation}
The attention path supplies content-dependent access to recent tokens and fixed global anchors. In V2, the projected Q/K/V vectors are eventized a second time. The selected event attention path uses per-head distance penalties from Attention with Linear Biases (ALiBi) \citep{press2022train} in place of rotary transformations, avoiding rotation of the discrete event vectors.

With zero-based positions and maximum lookback distance $w$, the visible set is
\[
 \mathcal K_i=\{j\in\mathbb Z:0\le j\le i,\quad i-j\le w\ \text{or}\ j<4\}.
\]
For head $h$, let $\bar q_i^{(h)},\bar k_j^{(h)},\bar v_j^{(h)}\in\mathbb R^{d_h}$ be the projected event vectors. The training-level attention definition is
\begin{align}
 s_{ij}^{(h)} &= \frac{(\bar q_i^{(h)})^{\mathsf T}\bar k_j^{(h)}}{\sqrt{d_h}}-m_h(i-j),
       &&j\in\mathcal K_i,\\
 a_{ij}^{(h)} &= \frac{\max(0,s_{ij}^{(h)})}
  {\sum_{k\in\mathcal K_i}\max(0,s_{ik}^{(h)})+\epsilon},
 & o_i^{(h)}&=\sum_{j\in\mathcal K_i}a_{ij}^{(h)}\bar v_j^{(h)}.
 \label{eq:attention}
\end{align}
The model has $n_{\rm heads}=12$ heads of dimension $d_h=64$, with slopes $m_h=2^{-8(h+1)/n_{\rm heads}}$ for $h=0,\ldots,n_{\rm heads}-1$ and $\epsilon=10^{-6}$. The four anchors are the first four sequence positions, subject to the causal mask. Positions outside $\mathcal K_i$ receive zero attention weight; if every permitted score is non-positive, the attention output is zero. ReLU--$L_1$ normalization removes exponentiation, while scaling, accumulation, and output projection remain in the execution budget. The FPGA implements these operations with deterministic integer arithmetic: its zero-sum branch returns zero, and positive sums use specified integer division and rounding rather than evaluating the training-time $\epsilon$ expression.

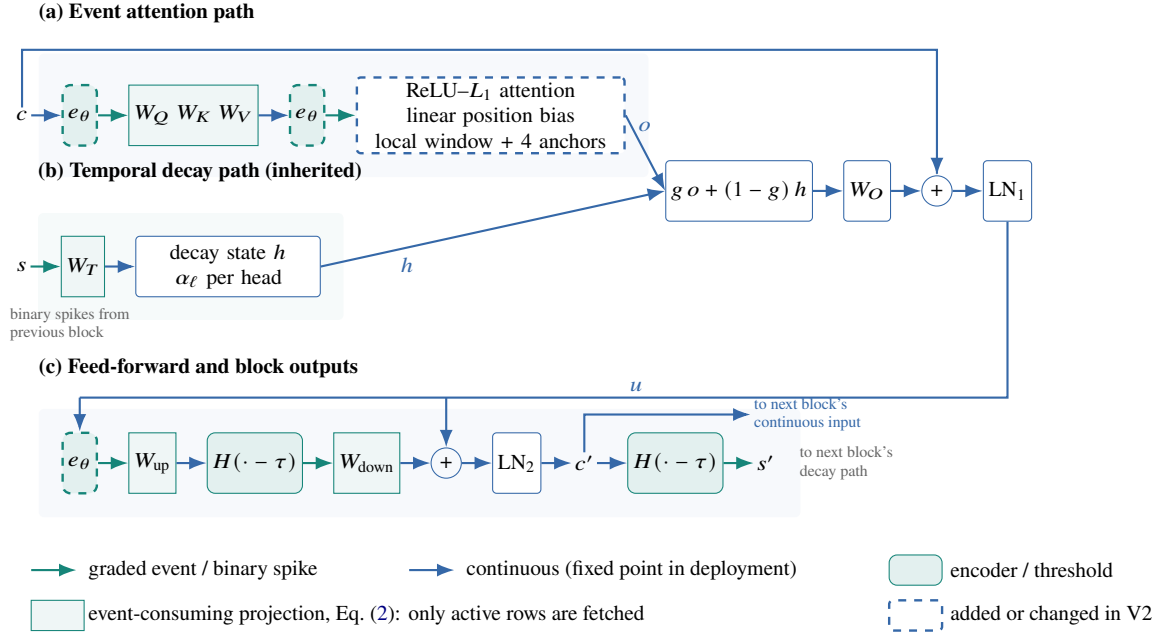
\begin{figure}[!htbp]
\centering
\begin{tikzpicture}[>=Latex,font=\scriptsize,
  every node/.style={inner sep=2pt},
  op/.style={draw=slblue,fill=white,rounded corners=1.5pt,align=center,minimum height=8mm},
  proj/.style={draw=slgreen,fill=slgreen!7,align=center,minimum height=8mm},
  enc/.style={draw=slgreen,fill=slgreen!12,rounded corners=3pt,align=center,minimum height=8mm},
  new/.style={dashed,line width=0.9pt},
  sum/.style={draw=slblue,circle,inner sep=0pt,minimum size=4mm},
  var/.style={inner sep=1pt},
  note/.style={inner sep=1pt,text=black!60,font=\tiny,align=left},
  ev/.style={->,thick,slgreen},
  ct/.style={->,thick,slblue},
  ctl/.style={thick,slblue}]
% Row 1: attention path
\node[var] (cin) at (0,0) {$c$};
\node[enc,new,right=4mm of cin] (enc1) {$e_\theta$};
\node[proj,right=4mm of enc1] (qkv) {$W_Q\;W_K\;W_V$};
\node[enc,new,right=4mm of qkv] (enc2) {$e_\theta$};
\node[op,new,right=4mm of enc2,text width=34mm] (attn) {ReLU--$L_1$ attention\\linear position bias\\local window $+$ 4 anchors};
% Row 2: decay path
\node[var] (sin) at (0,-20mm) {$s$};
\node[proj,right=4mm of sin] (wt) {$W_T$};
\node[op,right=4mm of wt,text width=23mm] (dec) {decay state $h$\\$\alpha_\ell$ per head};
\node[note,below=4mm of sin,anchor=north west,xshift=-2mm] {binary spikes from\\previous block};
% Fusion chain
\node[op,text width=18mm] (gate) at ($(attn.east)+(15mm,-10mm)$) {$g\,o+(1-g)\,h$};
\node[op,right=4mm of gate] (wo) {$W_O$};
\node[sum,right=4mm of wo] (add1) {$+$};
\node[op,right=4mm of add1] (ln1) {LN$_1$};
% Row 3: feed-forward path
\node[enc,new] (enc3) at ($(enc1.center)+(0,-46mm)$) {$e_\theta$};
\node[proj,right=4mm of enc3] (wup) {$W_{\rm up}$};
\node[enc,right=4mm of wup] (thr1) {$H(\cdot-\tau)$};
\node[proj,right=4mm of thr1] (wdown) {$W_{\rm down}$};
\node[sum,right=4mm of wdown] (add2) {$+$};
\node[op,right=4mm of add2] (ln2) {LN$_2$};
\node[var,right=4mm of ln2] (cout) {$c'$};
\node[enc,right=4mm of cout] (thr2) {$H(\cdot-\tau)$};
\node[var,right=4mm of thr2] (sout) {$s'$};
\node[note,right=2.5mm of sout] {to next block's\\decay path};
% Edges
\draw[ct] (cin)--(enc1);
\draw[ev] (enc1)--(qkv);
\draw[ct] (qkv)--(enc2);
\draw[ev] (enc2)--(attn);
\draw[ct] (attn.east)-- node[above,pos=0.3,xshift=1mm]{$o$} (gate.west);
\draw[ev] (sin)--(wt);
\draw[ct] (wt)--(dec);
\draw[ct] (dec.east)-- node[below,pos=0.25,yshift=-0.5mm]{$h$} (gate.west);
\draw[ct] (gate)--(wo);
\draw[ct] (wo)--(add1);
\draw[ct] (add1)--(ln1);
\draw[ctl] (cin.north) |- ($(attn.north)+(0,3.5mm)$) -| ($(add1.north)+(0,0.2mm)$);
\draw[ct] ($(add1.north)+(0,0.6mm)$)--(add1.north);
\coordinate (ubus) at ($(enc3.north)+(0,4.5mm)$);
\draw[ctl] (ln1.south) |- node[pos=0.7,above]{$u$} (ubus);
\draw[ct] (ubus)--(enc3.north);
\draw[ct] (add2.north |- ubus)--(add2.north);
\draw[ev] (enc3)--(wup);
\draw[ct] (wup)--(thr1);
\draw[ev] (thr1)--(wdown);
\draw[ct] (wdown)--(add2);
\draw[ct] (add2)--(ln2);
\draw[ct] (ln2)--(cout);
\draw[ct] (cout)--(thr2);
\draw[ct] (cout.north) -- ++(0,5mm) -- ++(22mm,0) node[right,note,text=slblue] {to next block's\\continuous input};
\draw[ev] (thr2)--(sout);
\begin{scope}[on background layer]
\node[fill=slblue!4,draw=none,rounded corners=2pt,fit=(enc1)(attn),inner xsep=3mm,inner ysep=3mm] (layer0) {};
\node[fill=slgreen!4,draw=none,rounded corners=2pt,fit=(wt)(dec),inner xsep=3mm,inner ysep=3mm] (layer1) {};
\node[fill=slblue!4,draw=none,rounded corners=2pt,fit=(enc3)(sout),inner xsep=3mm,inner ysep=3mm] (layer2) {};
\end{scope}
\node[anchor=south west,font=\scriptsize\bfseries,inner sep=0pt] at ($(layer0.north west)+(0,4mm)$) {(a) Event attention path};
\node[anchor=south west,font=\scriptsize\bfseries,inner sep=0pt] at ($(layer1.north west)+(0,4mm)$) {(b) Temporal decay path (inherited)};
\node[anchor=south west,font=\scriptsize\bfseries,inner sep=0pt] at ($(layer2.north west)+(0,4mm)$) {(c) Feed-forward and block outputs};
% Legend
\begin{scope}[shift={($(enc3.south west)+(-4mm,-10mm)$)},every node/.style={inner sep=1.5pt}]
\draw[ev] (0,0)--(6mm,0); \node[anchor=west] at (7mm,0) {graded event / binary spike};
\draw[ct] (50mm,0)--(56mm,0); \node[anchor=west] at (57mm,0) {continuous (fixed point in deployment)};
\node[enc,minimum height=4mm,minimum width=7mm] at (117mm,0) {}; \node[anchor=west] at (121mm,0) {encoder / threshold};
\node[proj,minimum height=4mm,minimum width=7mm] at (3.5mm,-6mm) {}; \node[anchor=west] at (7mm,-6mm) {event-consuming projection, Eq.~\eqref{eq:gather}: only active rows are fetched};
\node[op,new,minimum height=4mm,minimum width=7mm] at (117mm,-6mm) {}; \node[anchor=west] at (121mm,-6mm) {added or changed in V2};
\end{scope}
\end{tikzpicture}
\caption{One hybrid neuromorphic V2 block, corresponding to Equations~\eqref{eq:decay}--\eqref{eq:block}. Green edges carry integer events or binary spikes into the shaded projections, which fetch only the rows selected by nonzero inputs; blue edges carry continuous values that both deployments represent in fixed point. Relative to V1, V2 adds the three $e_\theta$ encoders, so that $W_Q$, $W_K$, $W_V$, and $W_{\rm up}$ now consume events, and replaces softmax attention with the ReLU--$L_1$ event attention path. $W_T$ and $W_{\rm down}$ keep V1's binary-spike inputs; the residual stream, $W_O$, and the layer norms remain continuous. The U50C FPGA prototype and ARM implementation execute this same integer graph through different dataflows.}
\label{fig:block}
\end{figure}

Let $o$ concatenate the head outputs $o_i^{(h)}$, $g$ be the layer gate, and $c$ the continuous block input. Omitting dropout and layer/position indices, fusion and the feed-forward path are
\begin{align}
 u &= \operatorname{LN}_1\!\left(c+W_O[g\,o+(1-g)h]\right),\\
 c'&=\operatorname{LN}_2\!\left(u+W_{\rm down}
 H(W_{\rm up}e_\theta(u)-\tau)\right),\qquad s'=H(c'-\tau).
 \label{eq:block}
\end{align}
Here $\operatorname{LN}$ denotes layer normalization, $H$ is the unit-step function with $H(0)=1$, $\tau$ is the spike threshold, and $g=\sigma(\gamma_\ell)$ is one learned scalar gate per layer. The threshold operations use surrogate gradients in training; they are not additional stateful LIF neurons. The final vocabulary projection and context-conditioned decoding head remain part of the full graph. \Cref{fig:block} draws one block: Equations~\eqref{eq:decay}--\eqref{eq:block} retain V1's two temporal paths and continuous residual transport while adding event-coded projection inputs and event attention. The binary spikes $s$ entering the decay projection come from the previous block's threshold output $s'$, or from the LIF input encoder in the first block.

\subsection{Configuration and hybrid computation}
\begin{table}[!htbp]
\centering\small
\caption{Architecture and training settings. V1 and V2 use different training corpora and tokenizers, so their absolute perplexities cannot directly measure quality changes between versions.}
\label{tab:architecture}
\begin{tabularx}{\linewidth}{@{}p{36mm}YY@{}}
\toprule
\bfseries Property & \bfseries SymbolicLight V1 & \bfseries SymbolicLight V2 \\
\midrule
Hybrid foundation & Spike-gated dual paths with a continuous residual & Inherited dual paths and residual; broader event computation \\
Core layout & 12 layers, width 768, FFN 4096 & Same core dimensions; 194,016,925 parameters \\
Event representation & Binary spikes; LIF input encoder & Binary inherited sites plus graded signed events \\
Event-consuming projections & Decay and FFN down & Additionally FFN up and Q/K/V inputs \\
Attention & Local softmax with continuous Q/K/V & Eventized Q/K/V; ReLU--$L_1$; local window plus four anchors \\
Position mechanism & Rotary embeddings & Linear position bias \\
Residual and output head & Continuous & Continuous in the model; fixed point in deployment \\
Training settings & 3B-token V1 protocol & Open-data 4B-token training; 48K tokenizer \\
Hardware evaluated here & No new V1 hardware measurements & INT8-weight ARM inference and an end-to-end U50C FPGA prototype \\
\bottomrule
\end{tabularx}
\end{table}
V2's hybrid computation assigns complementary roles to events and continuous-valued processing. Binary spikes and graded signed events select projection contributions; the recurrent decay state accumulates temporal information, local attention retrieves context, and the continuous residual carries features across blocks. Normalization, the attention output projection, and the output head retain continuous-valued operations, represented in fixed point during deployment. This architecture is therefore not an exclusively event-based spiking neural network (SNN). The U50C is the FPGA implementation platform for the hybrid architecture.

Training uses lookback distance $w=256$: at most 257 local positions including the current token, plus visible anchors outside that range. Deployment uses $w=475$, allowing at most 476 local positions plus four anchors, or 480 KV slots; overlapping positions are counted once. The FPGA stores these entries in a 480-slot circular buffer, reusing slots as the local window advances while preserving the anchors; ARM applies the same visibility mask. The short p32/n128 experiment remains within the training window; longer-prefix results evaluate the extended deployment configuration.

\section{From ARM Execution to Sparse FPGA Hardware}
\label{sec:hardware}
\subsection{Executable event projections on ARM}
The CPU runtime executes V2 on four Cortex-A76 cores of the RK3588. OpenMP manages four parallel threads, configured to sleep rather than spin while waiting. Weights are stored as signed 8-bit integers (INT8) and packaged once for deployment. At event-consuming projections, the runtime lists the nonzero inputs, reorders their contributions, and uses ARM NEON vector instructions for integer multiply-accumulate operations; dense projections use the dot-product path. The projection loop in Equation~\eqref{eq:gather} visits only nonzero contributions. Continuous operators and attention are evaluated separately.

The runtime's logical projection-MAC counter reports executed-to-dense-enumeration ratios of 0.482, 0.435, and 0.403 for p32/n128, p128/n128, and p256/n128, corresponding to 51.8--59.7\% fewer projection contributions. The counters cover matrix projections and exclude attention dot/value loops, normalization, and recurrence. They quantify skipped projection work; the energy measurements evaluate the complete implementation, without a sparse-off CPU control.

\subsection{Integer semantics and autonomous execution}
The Alveo U50C FPGA runs at 175\,MHz with INT8 weights, fixed-point activations and KV state, integer event payloads, wide accumulators, and deterministic requantization. A software deployment reference specifies its arithmetic, state transitions, and ring behavior. Bit-exactness denotes agreement with this integer reference, rather than with the FP32 training checkpoint.

A common host stack handles tokenization, prompts, session control, and detokenization. Weights are uploaded once and remain resident. The device generates tokens autonomously, maintains recurrent and KV state, and returns greedy token IDs. The software model is used only for validation; measured FPGA generation receives neither activations nor gathered weights from CPU reference execution.

Two operating modes expose different costs. A continuous start generates a fixed number of tokens on the device and amortizes host control. Interactive operation resumes once per token and exposes that token immediately. The latter supports multi-turn conversation and exact end-of-sequence (EOS) stopping, but includes a host round trip for every token. Neither mode batches prompt ingestion: prefill advances one token per model pass.

\subsection{From dense enumeration to active-row gathering}
We compare three successive implementations built from the same hardware design and trained model weights, named here by their distinguishing mechanism. The \impldense{} implementation keeps weights resident in HBM but enumerates dense weight rows at event sites while preserving the event arithmetic numerically. The \implgather{} implementation instead uses active event indices to fetch only the contributing rows in Equation~\eqref{eq:gather}. This upgrades the relationship between model sparsity and memory traffic: zeros can suppress physical reads rather than merely multiply fetched weights by zero.

Simulation accounts for 52\% fewer weight bytes than dense enumeration on its tested workload; this measures simulated traffic, not board energy. The corresponding FPGA build reduces per-pass cycles by 13.9\% relative to \impldense{}. Non-sparse operators, memory service, and control still contribute to token latency, limiting the whole-pass gain. These comparisons evaluate successive implementations; they do not switch sparsity on and off within one bitstream.

\subsection{Loading only populated KV state}
The \implkv{} implementation retains the \implgather{} data plane and reduces unnecessary KV loading at early session positions. Before the ring is full, loading its entire capacity transfers state that cannot yet contribute to attention. The \implkv{} build loads only the populated state required by the affected layers, following the software reference's cache-indexing rules. The final layer still loads the entire ring so that state read back for validation matches the reference. After the ring is full, this opportunity disappears.

The workload dependence makes this change testable. Continuous decode throughput improves from 544.2 to 643.2~\tps{} at p32/n128, while p480/n128 changes from 400.5 to 399.7~\tps{}. The near-zero change at the full-ring workload is consistent with the targeted removal of short-context state traffic. 

\subsection{Why these changes can lower energy}
Consider one interval of duration $T>0$ producing $N>0$ tokens. Let $R=N/T$ and $P_{\rm active}=T^{-1}\int_0^T P(t)\,\mathrm dt$, where $P(t)$ is instantaneous power. With mean loaded-idle power $P_{\rm idle}$ as the baseline, gross energy per generated token decomposes as
\begin{equation}
 E_{\rm gross}=\frac{P_{\rm active}}{R}
 =\frac{P_{\rm idle}}{R}+\frac{P_{\rm active}-P_{\rm idle}}{R}
 = E_{\rm idle\ share}+E_{\rm incremental}.
 \label{eq:energy}
\end{equation}
Power in watts divided by throughput in tokens/s gives J/token. The two terms allocate the idle baseline and the increment above it over the same duration. For complete requests, $T$ includes prefill and decode. Dividing mean whole-run power by decode-only throughput instead yields an estimate, rather than separately integrated decode energy. Sparse row gathering can reduce work and memory traffic at eligible projections. Partial KV loading can shorten execution even when the skipped intervals consume little power beyond the platform idle level. Increased throughput then amortizes the idle share over more tokens. Resident weights and an on-device token loop also remove repeated loading and host orchestration from steady-state generation. The measurements below separate these effects where the evidence permits.

\begin{figure}[!htbp]
\centering
\includegraphics[width=\linewidth]{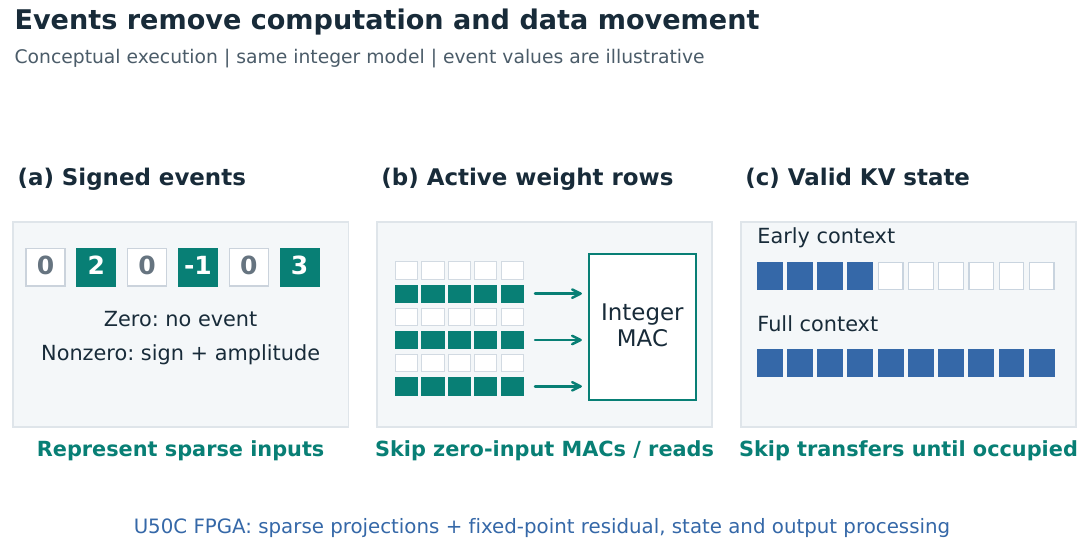}
\caption{Representation-to-execution mapping. Event values at left are illustrative; nonzero values retain sign and amplitude. Active inputs select weight rows, omitting zero-input MACs and reads. Valid-state KV loading avoids unused transfers before the ring fills; this state optimization can also benefit dense projections. Continuous residuals, recurrent state, and output processing complete the hybrid inference graph, implemented digitally in fixed point in the U50C FPGA prototype.}
\label{fig:execution_map}
\end{figure}

\section{Evaluation Protocol}
\label{sec:protocol}
\subsection{Workloads and matched model settings}
We write p$P$/n$N$ for a $P$-token prefix and $N$ generated tokens. Prefill processes the input sequence; decode generates output tokens one at a time. The requests in the throughput and energy tables use one sequence, greedy decoding, and $N=128$. The ROCK~5T matrix uses $P\in\{32,128,256\}$; the FPGA also includes p480/n128. The numerator counts \emph{generated tokens only}, including when the denominator contains prefill:
\begin{align}
 R_{\rm decode}&=N/T_{\rm decode},\\
 R_{\rm request}&=N/(T_{\rm prefill}+T_{\rm decode}).
 \label{eq:throughput}
\end{align}
FPGA decode time is the host-observed duration of the complete device generation start, including control overhead. Its interactive mode includes one host resume per token. Device passes/s, derived from board cycles, is a separate diagnostic.

The ARM and FPGA V2 measurements use the same checkpoint, integer deployment arithmetic, input token vectors, and attention-window parameter of 475. The CPU performs one additional forward step using the final selected output, whereas the FPGA workload stops after selecting output 128. Generated-token counts and reference output sequences therefore align, but the reported CPU time includes that additional pass. The training configuration and older GPU benchmark use a window of 256, as discussed in \Cref{sec:arch,app:gpu}.

\paragraph{Energy boundaries.}
We report gross and loaded-idle-subtracted incremental energy for ARM and FPGA side by side, consistently in J per generated token, while distinguishing AC-input integration from DC-card estimates. ARM energy is an integrated measurement: a smart plug at the board adapter's AC input is integrated over complete requests and divided by generated tokens, so it includes the whole board, its adapter, and prefill. FPGA energy is a DC-card estimate: mean card power read through the Xilinx Runtime (XRT) over active runs, which excludes the host and power-supply losses, divided by decode or request throughput. The supplementary GPU energy in \Cref{app:gpu} is the analogous board-sensor estimate read through the NVIDIA Management Library (NVML). The FPGA/GPU card-sensor ratios in \Cref{sec:gpu_energy} compare the recorded deployments, whose two sensors are not cross-calibrated. Where a loaded-idle power is available, Equation~\eqref{eq:energy} further splits the platform's gross energy into an idle share and an incremental term.

\subsection{ROCK 5T measurement and same-board comparison}
The V2 remeasurement uses the unchanged V2 model package, CPU binary, and input files on a 16\,GB ROCK~5T under Debian 12. Execution uses four A76 cores and four OpenMP threads, with no NPU offload. Each of the three prefix lengths has three interleaved rounds. A round comprises warmup, 60 seconds of loaded idle, repeated complete requests for at least 60 seconds, and a further 60 seconds of loaded idle. The first five seconds of each idle interval are excluded from its power integration. All nine runs are included, totaling 183 timed requests. Each metric is calculated per round and then independently reduced to the median; values in different table columns need not reproduce the ratios of any single round.

The smart plug supplies integer-watt readings at approximately 4\,Hz. Boundary-interpolated trapezoidal integration gives active-window energy, which is divided by generated-token count to obtain gross J/token. Incremental energy subtracts the mean pre- and post-run loaded-idle power over the same active interval. The measurement excludes the separately powered external cooler. The plug has no independent calibration or second-meter cross-check, so its integer-watt resolution does not establish absolute accuracy.

Gross-energy inter-round spreads, defined as $(\max-\min)/\mathrm{median}$, fall below the prespecified 5\% threshold. An earlier p32 series had 6.3\% spread after a fourth round; the unchanged model and runtime mean that differences between these series cannot be attributed to optimization. The p128 incremental-energy spread is 6.32\%, limiting comparisons of small incremental differences. The latest runs reach at most 73.0\,$^\circ$C with no cpufreq cooling action. Two start-of-load telemetry samples show a lower frequency; all other recorded active samples show 2.256\,GHz on the big cores. The 1\,Hz telemetry cannot establish uninterrupted constant frequency.

The comparison paths were measured earlier on the same board and day: Qwen2.5-0.5B and SmolLM2-135M with the RKLLM inference runtime using 8-bit weights and activations (W8A8) on the NPU, and both models plus LFM2.5-350M with the llama.cpp inference runtime using its Q8\_0 weight-quantization format on four A76 cores. NPU paths use the three-core NPU with big-core support. Each comparison cell has three rounds under the same complete-request energy protocol. These models differ in parameter count, tokenizer, input sequence, quantization format, and runtime. Equal prefix/output token counts define a deployment benchmark, not equal text content or equal task quality.

The earlier llama.cpp CPU paths operated at approximately 87.7--90.3\,$^\circ$C, warmer than V2 and the NPU paths. A postmeasurement protocol amendment retained runs above 85\,$^\circ$C when no cpufreq cooling action occurred. These thermal conditions differ across paths, and the latest remeasurement covers V2 alone.

\subsection{FPGA energy and correctness}
The focused U50C rerun uses three rounds of at least 60 seconds per shape, with XRT/\texttt{xbutil} card telemetry at approximately 0.5\,Hz. Mean active-run power divided by decode throughput gives the decode-energy estimate; dividing by whole-request throughput gives the corresponding request estimate. Decode energy is the median of per-round power/throughput ratios; request energy divides the independently aggregated median power by median request throughput. Power sampling spans both prefill and decode.

For the complete-request ARM comparison, we additionally reanalyze the nine retained FPGA runs. Each run provides a 10-second loaded-idle measurement before activity. We compute $(\bar P_{\rm run}-\bar P_{\rm idle})/R_{\rm request}$ within each run and report the three-round median, using the same run power and request throughput as the gross-energy calculation. This is a secondary analysis of existing records, with no new measurements. Its pre-run-only idle interval is shorter than the two 60-second intervals on ARM. Subtracting idle energy does not remove differences in device coverage, supply losses, or sensors.

The independent FPGA idle-split run measures loaded idle before and after activity with weights retained. Active power is calculated from non-warmup execution windows after the initial idle interval. Interactive energy combines timing with power sampling disabled and a separate run with power sampling enabled; it is reported as a split-run estimate.

ARM validation finds full-logit and state-hash agreement on 264 reference tokens under the original 256-window configuration. At the deployed 475-window setting, p32/p128/p256 sequences match the U50C reference for 128 generated tokens per shape. All nine saved final-request sequences from the remeasurement also match the reference sequences; outputs were not saved for all 183 requests. FPGA validation fixes the model and bitstream identities and checks 14 scenarios, including three rounds of continuous and one-token-resume workloads. A 30-minute continuous stability test completes 2,263 conversations and 27,148 turns, including 1,131 ring-wrap conversations, with 47 exact state probes and approximately 0.014\% drift in time per output token (TPOT). These tests check integer execution and session stability; language quality is evaluated separately.

\section{Energy and Execution-Efficiency Evaluation}
\label{sec:results}
\subsection{Same-checkpoint energy and throughput progression}
\begin{figure}[!htbp]
\centering
\includegraphics[width=\linewidth]{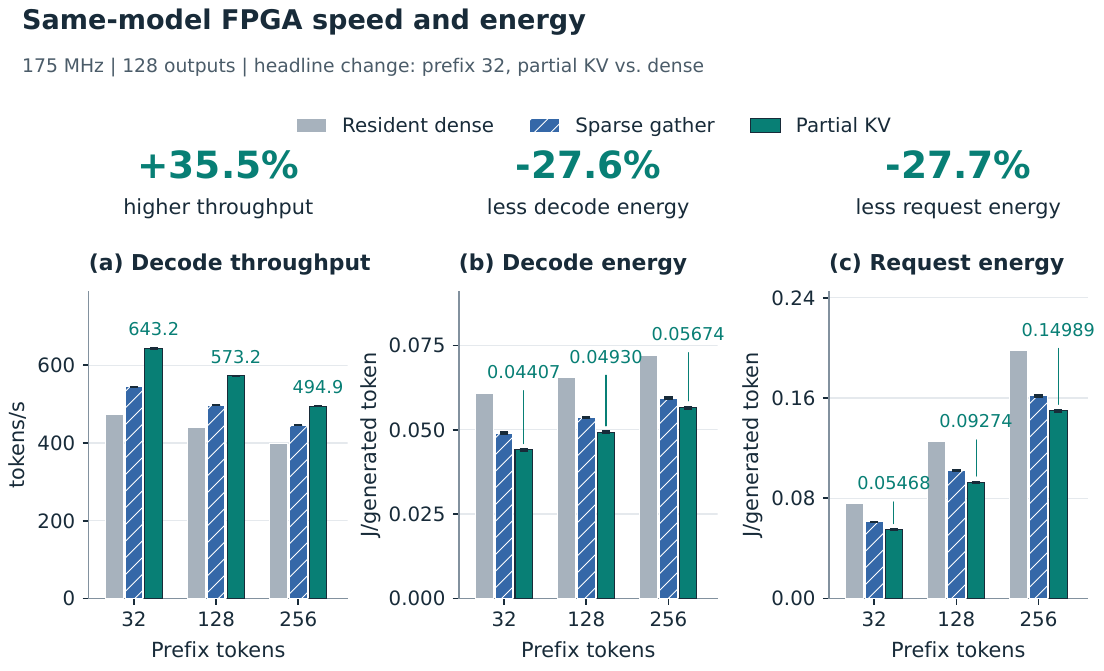}
\caption{Same-checkpoint FPGA builds at 175 MHz. Panels show decode throughput, decode energy, and complete-request energy; each request generates 128 tokens. Throughput and decode energy are three-round medians; request energy is median power divided by median request throughput. Error bars on the two sparse builds span per-round minima and maxima; the dense baseline shows only the available summary values. Energy per generated token is estimated from card power and throughput; thermal conditions differ.}
\label{fig:fpga_progression}
\end{figure}

Two FPGA implementation upgrades improve both throughput and gross energy per generated token with the model weights held fixed (\Cref{tab:evolution,fig:fpga_progression}). At p32/n128, \impldense{}, \implgather{}, and \implkv{} reach 474.6, 544.2, and 643.2 decode tokens/s, respectively, a cumulative 35.5\% increase. Estimated gross decode energy falls from 0.06087 to 0.04407\,J/generated token (27.6\%). Including prefill, short-request energy falls from 0.07561 to 0.05468\,J/generated token (27.7\%).

\begin{table}[!htbp]
\centering\small
\caption{Three FPGA implementations of the same checkpoint, prefix 32 and 128 generated tokens. Three rounds per implementation: power and throughput are separate medians, decode energy is the median of per-round ratios, and request energy is median power divided by median request throughput. Energy is in J/generated token. Thermal conditions differ across measurement campaigns. Bold identifies the partial-KV implementation.}
\label{tab:evolution}
\begin{tabular}{@{}lrrrr@{}}
\toprule
\bfseries Implementation & \bfseries Decode tok/s & \bfseries Card W & \bfseries Decode J/token & \bfseries Request J/token\\\midrule
\impldense{} & 474.6 & 28.89 & 0.06087 & 0.07561\\
\implgather{} & 544.2 & 26.69 & 0.04904 & 0.06108\\
\bfseries \implkv{} & \bfseries  643.2 & \bfseries  28.34 & \bfseries  0.04407 & \bfseries  0.05468\\
\bottomrule
\end{tabular}
\end{table}

At prefixes 128 and 256, complete-request energy falls by 25.9\% and 24.4\%, respectively. The reduction therefore persists when prompt ingestion is included. These builds preserve the weights and pass integer-reference validation. The changes quantify the combined implementation progression, without separately identifying thermal effects and every control change.

The two steps remove different work. Active-row gathering omits weight accesses for zero events; partial KV loading avoids moving unused state before the cache fills. Faster sparse projections expose KV-loading waits that were previously overlapped with computation, so the second change releases further execution benefit. The p32 decode rate rises from 544.2 to 643.2 tokens/s, whereas p480 changes from 400.5 to 399.7, consistent with the disappearance of avoidable transfers at a full ring.

\begin{table}[!htbp]
\centering\small
\caption{Mechanisms and evidence levels. Denominators and workloads differ across rows; the percentages are not interchangeable energy reductions.}
\label{tab:evidence_chain}
\begin{tabularx}{\linewidth}{@{}p{29mm}YY@{}}
\toprule
\bfseries Evidence level & \bfseries Result & \bfseries Scope\\\midrule
ARM projections & 51.8--59.7\% of MAC terms skipped & Nonzero-input loops, three prefixes\\
FPGA weight reads & 52\% fewer weight bytes & Simulation on the tested vectors\\
FPGA end-to-end & 24.4--27.7\% less request energy & Same-checkpoint builds, card estimate\\
\bottomrule
\end{tabularx}
\end{table}

\Cref{tab:evidence_chain} connects representational opportunity, execution behavior, and board measurements. ARM counters show 51.8\%, 56.5\%, and 59.7\% fewer projection MAC terms at the three prefixes. FPGA simulation records 52\% fewer weight bytes on its tested vectors. An earlier full-model prefill trace counts 159,850,985 dense-equivalent and 80,909,678 active MACs/token, a 49.38\% reduction. These counts explain removable work; measured card energy evaluates the resulting implementation.

\subsection{Gross energy and the loaded-idle share}
\begin{figure}[!htbp]
\centering
\includegraphics[width=\linewidth]{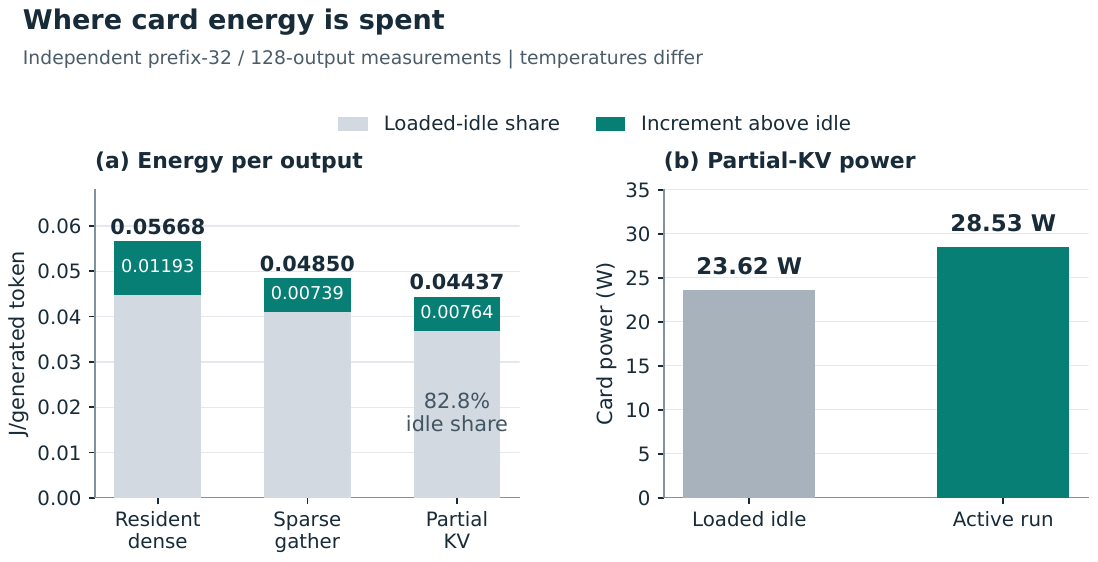}
\caption{Independent p32/n128 continuous idle-split runs. Left: gross card energy separates into a loaded-idle share and activity above idle; top labels are totals and dark-segment labels are increments, all in J/generated token. Right: loaded-idle and active power for partial KV. Thermal conditions differ across builds, so the chart is not an isolated dynamic-energy ablation.}
\label{fig:energy_split}
\end{figure}

The independent \implkv{} idle-split experiment generates 31,104 tokens over 243 repetitions, with all recorded output sequences exact to the reference. It measures 643.03~\tps{}, 28.534\,W over active-run windows, and 23.623\,W loaded idle. Equation~\eqref{eq:energy} gives
\begin{equation}
 E_{\rm gross}\approx0.04437\;\mathrm{J/token},\quad
 E_{\rm incremental}\approx0.00764\;\mathrm{J/token},\quad
 E_{\rm idle\ share}\approx0.03674\;\mathrm{J/token}.
\end{equation}
As shown in \Cref{fig:energy_split}, about 82.8\% of gross card energy in this experiment is the loaded-idle share. Faster generation therefore reduces the idle energy allocated to each token, even if active power changes little. The approximately 4.91\,W increment measures activity above loaded idle on the FPGA card.

The \implkv{} idle split followed an extended test campaign: FPGA temperature increased from 74 to 77\,$^\circ$C and HBM from 69 to 72\,$^\circ$C. The earlier \implgather{} run used a cooler card with approximately 22.4\,W idle. That run measured about 0.049\,J/token gross and 0.0074\,J/token incremental energy; the \implkv{} run measured 0.0444 and 0.0076, respectively. A comparison of these runs under different thermal conditions shows lower gross energy and similar incremental energy for \implkv{}. It does not isolate dynamic-energy savings or attribute the improvement solely to fewer arithmetic operations.

\subsection{The same V2 deployment on ARM and FPGA}
\begin{figure}[htbp]
\centering
\includegraphics[width=\linewidth]{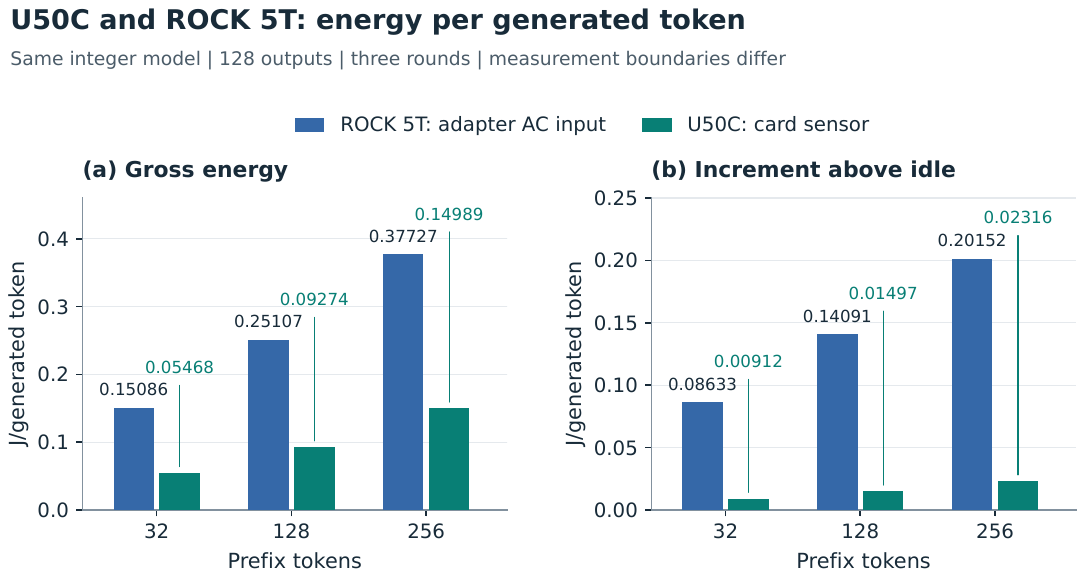}
\caption{Complete-request energy for the same integer model, generating 128 tokens. Left: gross energy; right: incremental energy above loaded idle, both in J/generated token. ROCK 5T integrates adapter AC input; U50C estimates card energy from power and request throughput. Each prefix has three rounds, aggregated as in \Cref{tab:arm_fpga}; panel scales differ. Subtracting idle does not equalize the two measurement boundaries.}
\label{fig:arm_fpga_energy}
\end{figure}

\begin{table}[htbp]
\centering\small
\caption{Complete-request throughput and energy for the same V2 weights, integer arithmetic, inputs, and 475-window setting; each request generates 128 tokens. ARM includes an extra forward step after selecting the final output. ROCK 5T measures AC input including its adapter; U50C estimates card energy excluding the host. Incremental energy subtracts the same-run loaded-idle baseline. FPGA increments are derived from retained records; aggregation is specified in \Cref{sec:protocol}.}
\label{tab:arm_fpga}
\begin{tabular}{@{}lrrrrrr@{}}
\toprule
 & \multicolumn{3}{c}{\bfseries ROCK 5T: AC input} & \multicolumn{3}{c}{\bfseries U50C: card sensor}\\
\bfseries Prefix & \bfseries Request tok/s & \bfseries Gross & \bfseries Increment & \bfseries Request tok/s & \bfseries Gross & \bfseries Increment\\
 & & \bfseries J/token & \bfseries J/token & & \bfseries J/token & \bfseries J/token\\\midrule
32 & 65.4 & 0.15086 & 0.08633 & 518.4 & 0.05468 & 0.00912\\
128 & 37.5 & 0.25107 & 0.14091 & 304.8 & 0.09274 & 0.01497\\
256 & 23.7 & 0.37727 & 0.20152 & 187.3 & 0.14989 & 0.02316\\
\bottomrule
\end{tabular}
\end{table}
Both gross and incremental energy show lower generation cost on U50C within the measured boundaries (\Cref{tab:arm_fpga,fig:arm_fpga_energy}). At prefix 32, complete-request U50C energy is 0.05468\,J/generated token gross and 0.00912\,J/generated token above loaded idle, versus 0.15086 and 0.08633, respectively, on ROCK~5T. Both metrics retain this ordering at prefixes 128 and 256. U50C draws more active power but completes requests faster, reducing energy per generated token. These results compare the recorded deployment boundaries; they do not give a whole-system FPGA efficiency ratio including the host and supply losses.

Incremental energy is particularly sensitive to the idle baseline. Its three-round relative spreads on U50C are 2.16\%, 3.19\%, and 2.22\% at the three prefixes, versus 2.57\%, 6.32\%, and 3.00\% on ROCK~5T. Both platforms include prefill in this comparison. The U50C complete-request increment of 0.00912\,J/token therefore belongs to a different measurement from the independent decode-only increment of 0.00764\,J/token in the preceding subsection.

With the model and deployment window fixed, the dedicated implementation provides higher decode and complete-request throughput (\Cref{tab:arm_fpga}). At p32 the ARM runtime reaches 65.4 generated tokens/s per complete request, and the FPGA 518.4. The FPGA handles prompt tokens sequentially but processes projections and state through its resident data plane; p256 prompt ingestion takes 424.6\,ms, versus approximately 3.000\,s on ARM. This comparison measures the combined effect of hardware capacity, memory organization, and implementation.

\begin{table}[!htbp]
\centering\small
\caption{U50C \implkv{} FPGA, continuous decode, batch one, 128 generated tokens. Three-round medians of host-observed throughput and mean XRT card power; energy is the DC-card estimate of \Cref{sec:protocol}, obtained by dividing mean power by decode or request throughput.}
\label{tab:fpga}
\begin{tabular}{@{}lrrrrr@{}}
\toprule
\bfseries Prefix & \bfseries Decode \tps{} & \bfseries Request \tps{} & \bfseries Card W & \bfseries Decode J/token & \bfseries Request J/token\\
\midrule
32  & 643.2 & 518.4 & 28.34 & 0.04407 & 0.05468\\
128 & 573.2 & 304.8 & 28.26 & 0.04930 & 0.09274\\
256 & 494.9 & 187.3 & 28.08 & 0.05674 & 0.14989\\
480 & 399.7 & 103.3 & 27.60 & 0.06905 & 0.26730\\
\bottomrule
\end{tabular}
\end{table}
\Cref{tab:fpga} gives the FPGA energy estimates. Mean card power varies by less than 1\,W across prefixes, so longer context raises energy mainly by increasing token time. The p32 three-round throughput spread is 0.04\%, and an independent synchronized measurement gives 643.5~\tps{}. The p480 request values are derived from the measured request throughput and the same card power. Across the three-round series of both sparse builds, decode-energy spreads range from 0.47 to 1.72\%.

\begin{figure}[!htbp]
\centering
\includegraphics[width=\linewidth]{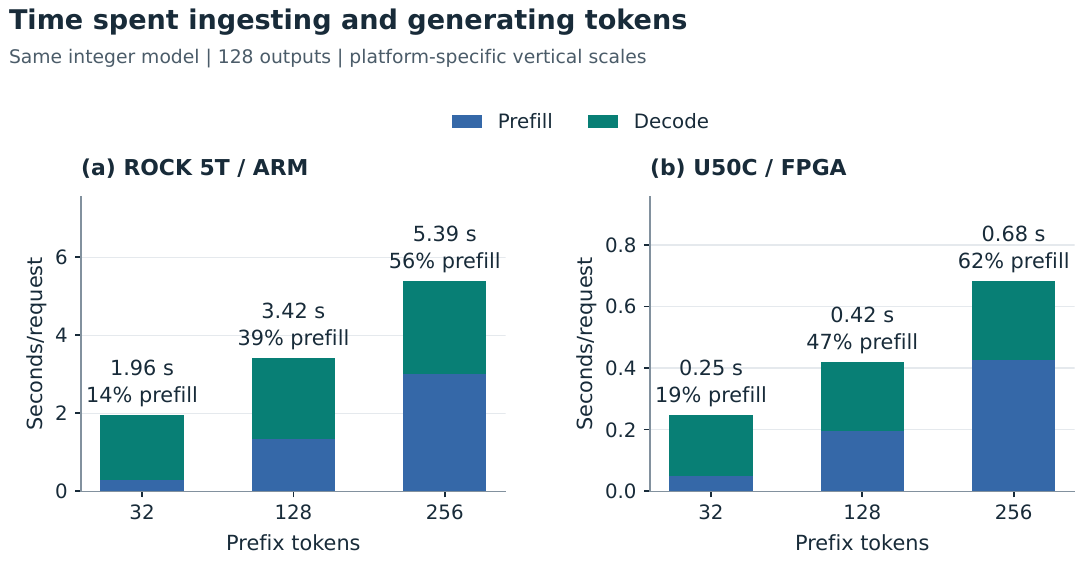}
\caption{Prompt ingestion and generation time for the same integer model, with 128 generated tokens. Blue denotes prefill and green decode; top labels give their summed time in seconds and the prefill share. Platform axes use different scales. Each phase uses its median, so the sum can differ slightly from the independently aggregated whole-request median.}
\label{fig:request_cost}
\end{figure}
\subsection{Interactive operation}
Interactive \implkv{} generation reaches 231.8~\tps{}, or 4.31\,ms/token, over ten 128-token turns with power sampling disabled. Ten separate turns with power sampling enabled measure 23.32\,W, giving a split-run gross estimate of 0.1006\,J/token. The active-minus-idle difference is too small to resolve reliably across these runs, so incremental interactive energy is not reported. This per-token-resume workflow includes a host round trip for each output, unlike continuous device-side decoding.

Separate board-cycle diagnostics report 519.66, 603.46, and 730.25 forward passes/s for the three builds; these exclude application overhead and are not interactive token rates. The dedicated one-token-resume test records 239.43 tokens/s, separately from the interactive campaign above.
\subsection{Card energy against the recorded GPU deployment}
\label{sec:gpu_energy}
\begin{figure}[!htbp]
\centering
\includegraphics[width=\linewidth]{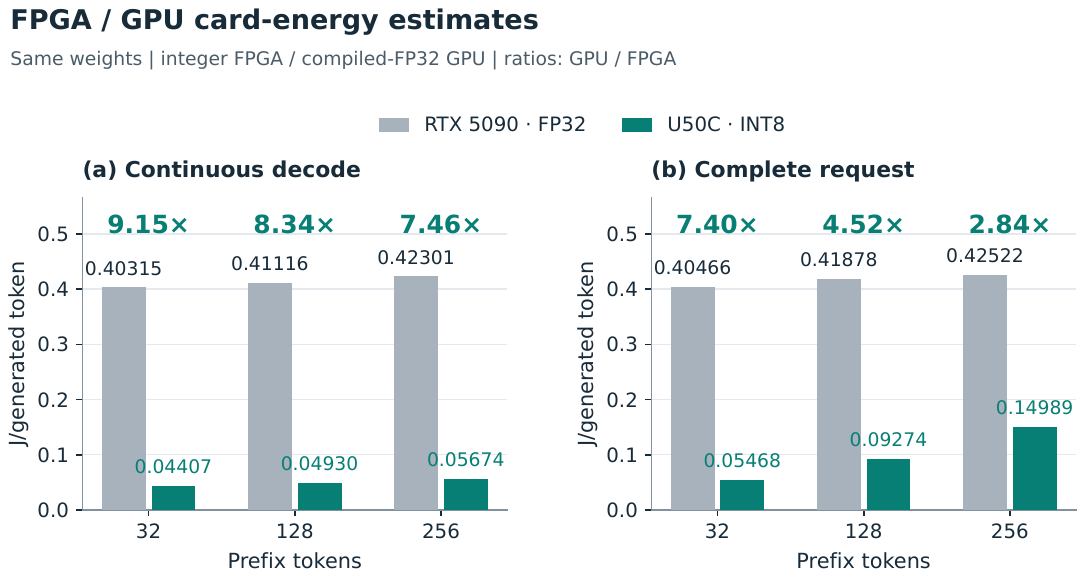}
\caption{Estimated card energy for two deployments of the same weights. Left: continuous decode; right: complete request including prefill, with 128 generated tokens. Bar labels are J/generated token; upper labels are GPU/FPGA energy ratios. U50C uses integer arithmetic and RTX 5090 compiled FP32. The comparison does not establish equal-quality efficiency or a GPU energy lower bound.}
\label{fig:gpu_energy}
\end{figure}

\Cref{fig:gpu_energy,tab:main} compares the integer FPGA deployment with the recorded RTX~5090 compiled-FP32 baseline for the same checkpoint. At p32/n128, FPGA continuous decode reaches 643.2 tokens/s and 0.04407\,J/generated token, versus 406.9 tokens/s and 0.40315\,J/generated token on the GPU. Here, compiled FP32 uses the PyTorch \texttt{torch.compile} optimizer while retaining FP32 arithmetic. The FPGA delivers 1.58$\times$ the throughput with 89.1\% lower estimated gross decode energy. Including prefill gives 0.05468 versus 0.40466\,J/generated token, a 7.40$\times$ energy ratio.

This comparison matches checkpoint, input tokens, and output length, but uses different arithmetic precisions and uncalibrated cross-device sensors; output identity between FP32 and integer arithmetic is not required. The short workload stays within both attention windows, while longer workloads differ in visible context. Compiled FP32 is the fastest recorded GPU path, not the minimum-energy configuration. Full protocols, precision screening, and the paired architecture experiment are in \Cref{app:gpu}.

\begin{table}[!htbp]
\centering\small
\caption{Continuous decode, 128 generated tokens, batch one. U50C: \implkv{} INT8-weight integer deployment, host-observed timing of physical hardware. RTX~5090: FP32 compiled path. Energy is gross board-sensor power/throughput. Three-round medians.}
\label{tab:main}
\begin{tabular}{@{}lrrrrrr@{}}
\toprule
 & \multicolumn{3}{c}{\bfseries U50C \implkv{}} & \multicolumn{2}{c}{\bfseries RTX 5090} & \bfseries Energy ratio\\
\bfseries Prefix & \bfseries \tps{} & \bfseries W & \bfseries J/token & \bfseries \tps{} & \bfseries J/token & \bfseries GPU/FPGA\\
\midrule
32  & 643.2 & 28.34 & 0.04407 & 406.9 & 0.40315 & 9.15$\times$\\
128 & 573.2 & 28.26 & 0.04930 & 411.2 & 0.41116 & 8.34$\times$\\
256 & 494.9 & 28.08 & 0.05674 & 406.3 & 0.42301 & 7.46$\times$\\
\bottomrule
\end{tabular}
\end{table}

\paragraph{Measured GPU low-precision alternatives.}
Lower precision can reduce GPU energy even when it does not increase throughput. \Cref{tab:gpu_precision} compares compiled FP32, bfloat16 (BF16), and weight-only INT8 on another RTX~5090, using the same V2 checkpoint. Each setting is measured twice at each of three prefix lengths (32, 128, and 256 tokens), always generating 128 tokens with batch one. Each run starts with three warmup requests, followed by at least 30 seconds of timed requests. Timing includes prefill and synchronizes once per generated sequence. For each setting, throughput and NVML card power are separately averaged with equal weight across the six runs. Reported energy is mean power divided by mean throughput; individual runs' energy ratios are not averaged.

\begin{table}[!htbp]
\centering\small
\caption{Selected GPU precision settings for the same V2 checkpoint, measured in one paired campaign on an RTX 5090. Complete requests with 128 generated tokens; prefixes of 32, 128, and 256 tokens, two rounds per setting. Power and throughput are arithmetic means over six runs. Energy is mean card power divided by mean request throughput, in J/generated token. The GPU and measurement protocol differ from those in \Cref{tab:main}.}
\label{tab:gpu_precision}
\begin{tabular}{@{}lrrr@{}}
\toprule
\bfseries GPU configuration & \bfseries Request tok/s & \bfseries Card W & \bfseries J/token\\\midrule
Compiled FP32 & 319.8 & 162.0 & 0.5065\\
Compiled BF16 & 312.1 & 141.1 & 0.4522\\
Compiled weight-only INT8 & 251.7 & 112.6 & 0.4475\\
\bottomrule
\end{tabular}
\end{table}
\begin{samepage}
Weight-only INT8 lowers gross card energy per generated token by 11.7\% relative to FP32 within this campaign, calculated before rounding, despite lower throughput. It quantizes weights and retains floating-point activation computation; its arithmetic therefore differs from the integer rules used throughout the FPGA inference implementation. These results establish a measured GPU energy improvement from low precision under the tested conditions. They do not establish an optimized-GPU energy lower bound, and their cross-prefix summaries cannot be substituted into the per-prefix FPGA/GPU ratios in \Cref{tab:main}. The 89.1\% headline reduction remains a comparison against the recorded compiled-FP32 deployment.
\end{samepage}

\subsection{Complete-request performance on ROCK 5T}
\begin{table}[!htbp]
\centering\small
\caption{V2 on four A76 cores of a ROCK~5T. Three-round medians; each request generates 128 tokens. Energy integrates complete requests at the board adapter's AC input and excludes external cooling. Each column is reduced independently; gross and incremental energy are in J/token.}
\label{tab:arm}
\begin{tabular}{@{}lrrrrrr@{}}
\toprule
 & & & & \multicolumn{2}{c}{\bfseries Energy (J/token)} & \bfseries \\
\bfseries Prefix & \bfseries Request \tps{} & \bfseries Active W & \bfseries Idle W & \bfseries Gross & \bfseries Incremental & \bfseries Request s\\
\midrule
32 & 65.4 & 9.80 & 4.22 & 0.151 & 0.086 & 1.957\\
128 & 37.5 & 9.41 & 4.12 & 0.251 & 0.141 & 3.416\\
256 & 23.7 & 8.94 & 4.18 & 0.377 & 0.202 & 5.392\\
\bottomrule
\end{tabular}
\end{table}
\Cref{tab:arm} shows complete local requests at approximately 9--10\,W. At p32, one 128-token response takes 1.96\,s and 19.3\,J. At p128 and p256, request energy rises to 32.1 and 48.3\,J despite slightly lower mean active power. Gross-energy spreads are 2.16\%, 3.15\%, and 0.40\%, respectively. Independent integration of the nine active and eighteen idle windows reproduces the reported energy estimates.

The prefix sensitivity is visible in measured phase timing. Median prefill time grows from 0.282 to 1.335 to 3.000\,s, while decode grows from 1.675 to 2.081 to 2.394\,s. Prompt ingestion accounts for about 14\% of the short request and 56\% of the long request. Decode throughput alone would therefore hide a substantial edge-runtime cost. The primary ARM energy result includes both phases.

\subsection{Same-board CPU and NPU deployment tradeoffs}
\begin{figure}[!htbp]
\centering
\includegraphics[width=\linewidth]{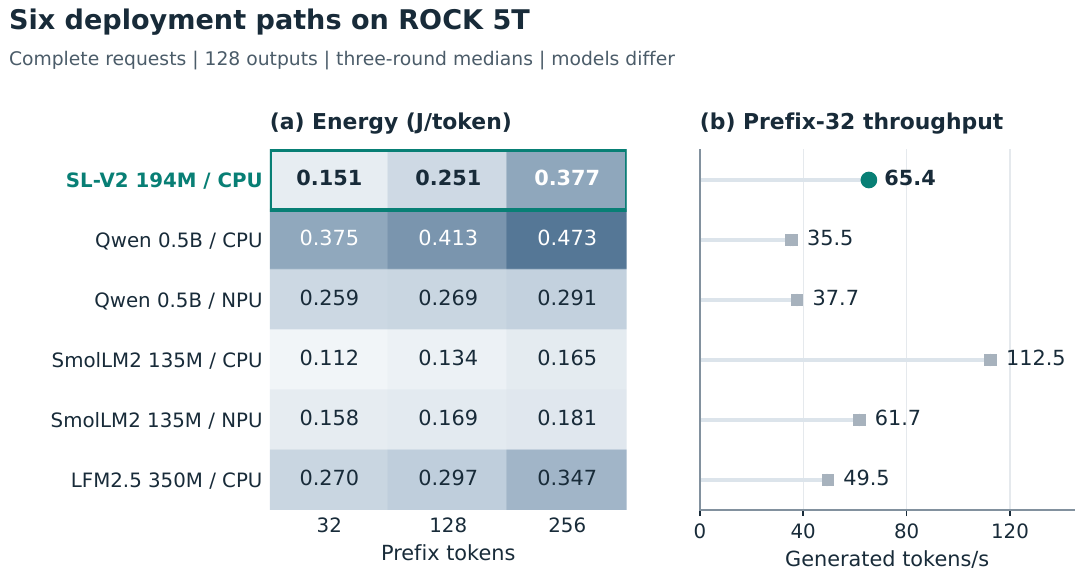}
\caption{All six ROCK 5T deployment paths, complete requests, 128 generated tokens. Left: AC-input energy across three prefixes, with lighter cells indicating lower energy. Right: request throughput at prefix 32; points are three-round medians and intervals span minima to maxima. V2 was remeasured without model or code changes; other deployments retain their earlier same-day measurements. Models, tokenizers, and runtimes differ.}
\label{fig:edge_comparison}
\end{figure}

\begin{table}[!htbp]
\centering\small
\caption{Same ROCK~5T, complete requests, 128 generated tokens. V2 was remeasured without model or code changes; the five comparison deployments retain their earlier same-day measurements. All rows report three-round medians. All energies are gross AC J/generated token. CPU means four A76 cores; NPU means RKLLM W8A8, with CPU support. Tokenizers, model sizes, inputs, and task quality differ. Bold identifies the V2 deployment.}
\label{tab:edge_comparison}
\begin{tabular}{@{}lrrrrrr@{}}
\toprule
 & \multicolumn{3}{c}{\bfseries\shortstack{Request \tps{}\\(higher is better)}} & \multicolumn{3}{c}{\bfseries\shortstack{J/token\\(lower is better)}}\\
\bfseries Model / path & \bfseries p32 & \bfseries p128 & \bfseries p256 & \bfseries p32 & \bfseries p128 & \bfseries p256\\
\midrule
\bfseries V2 194M / INT8 CPU & \bfseries  65.4 & \bfseries  37.5 & \bfseries  23.7 & \bfseries  0.151 & \bfseries  0.251 & \bfseries  0.377\\
Qwen 0.5B / Q8\_0 CPU & 35.5 & 31.8 & 27.4 & 0.375 & 0.413 & 0.473\\
Qwen 0.5B / W8A8 NPU & 37.7 & 36.1 & 34.0 & 0.259 & 0.269 & 0.291\\
SmolLM2 135M / Q8\_0 CPU & 112.5 & 93.7 & 74.8 & 0.112 & 0.134 & 0.165\\
SmolLM2 135M / W8A8 NPU & 61.7 & 57.4 & 53.6 & 0.158 & 0.169 & 0.181\\
LFM2.5 350M / Q8\_0 CPU & 49.5 & 42.9 & 36.6 & 0.270 & 0.297 & 0.347\\
\bottomrule
\end{tabular}
\end{table}
\Cref{tab:edge_comparison,fig:edge_comparison} compare the tested deployments. V2 uses 59.8\%, 39.2\%, and 20.2\% less gross energy than the Qwen CPU path at the three prefixes, respectively. At p32 it is also faster and uses less energy than Qwen NPU and LFM CPU. Its p32 result is close to SmolLM2 NPU: 65.4 versus 61.7~\tps{} and 0.151 versus 0.158\,J/token, with essentially identical incremental energy near 0.086\,J/token. The meter's uncertainty prevents interpreting this small gross-energy difference as a reliable efficiency advantage.

SmolLM2 CPU is faster and more energy-efficient at every prefix, despite its higher active board power. At p256, V2 uses more energy than both NPU paths and LFM CPU, and its request throughput falls more sharply with prefix length. The ARM implementation is therefore most competitive on short requests in this comparison. Model size, quantization, and runtime all differ across rows, so the comparison places V2 among deployable options; it does not isolate the effect of sparse execution (\Cref{sec:scope}).

\section{Discussion and Scope}
\label{sec:scope}
The main experiment evaluates the cost of executing a fixed checkpoint. Active-row gathering removes work associated with zero inputs, while valid-state KV loading reduces state transfer. Together with the measured loaded-idle share, these mechanisms explain the observed energy changes. Cross-model ARM results and different-precision FPGA/GPU results describe deployment tradeoffs; they do not isolate the energy contribution of event coding alone.

\paragraph{Extension to other dedicated hardware.}
The U50C prototype provides an implementation basis for dedicated hybrid neuromorphic V2 hardware. Its acceleration and energy-saving mechanisms are applicable across hardware implementations. Nonzero-event computation and weight fetching, valid-state KV transfers, resident weights, and on-device generation can also be implemented on other FPGAs or application-specific integrated circuits (ASICs). A target must support both graded signed events and the retained residual, normalization, and attention operations. The U50C measurements demonstrate that these mechanisms can be combined into a complete inference system, providing a design basis for other V2-compatible accelerators.

\begin{samepage}
A precise condition connects this portability to performance and energy. Consider a dedicated implementation $H$ and a reference implementation $B$ of the same V2 weights, arithmetic, inputs, context window, and output length. Let $R$ denote complete-request generation throughput, $\bar P$ the mean active power over that request interval, and $E$ gross energy per generated token. With matching timing definitions and device measurement boundaries, Equation~\eqref{eq:energy} yields
\begin{equation}
 \frac{E_H}{E_B}=\frac{\bar P_H/\bar P_B}{R_H/R_B},
 \qquad
 \frac{R_H}{R_B}>\max\!\left(1,\frac{\bar P_H}{\bar P_B}\right)
 \;\Longrightarrow\; R_H>R_B\ \text{and}\ E_H<E_B.
 \label{eq:hardware_generalization}
\end{equation}
\end{samepage}
Thus, increasing throughput by a larger factor than the change in active power delivers both faster generation and lower energy per generated token. This relation uses power and throughput from the same measurement interval; it does not directly combine different device boundaries or independently reduced table medians.

Whether another target meets this condition depends on the computation, memory traffic, and time saved relative to event-indexing, scheduling, and data-reordering overheads, as well as the capacity and bandwidth available to the dataflow. KV-transfer savings also depend on cache occupancy. The resulting inference is conditional but extends beyond the U50C: other V2-compatible dedicated implementations have a mechanism-supported route to faster, lower-energy generation, with the realized gain determined by hardware design and workload.

The deployed checkpoint has held-out FP32 perplexity 14.75 versus 12.50 for the same-budget dense control, approximately 18\% higher, and exhibits reduced accuracy on a key task suite. The same-weight implementation upgrades leave this model unchanged: they lower execution cost without establishing equal-quality efficiency. Full quality results, the attention variant, and quantization diagnostics are reported in \Cref{app:quality}.

The FPGA uses one sequence, a finite KV ring, and greedy output. On-board non-greedy sampling and full-logit readout are not implemented. The runtime disables early EOS in continuous multi-token mode because of a recorded state fault; per-token interaction supports exact stopping. ARM measurements exclude external cooling, and FPGA/GPU measurements exclude host and supply losses. Incremental FPGA energy is not used as an ASIC energy projection.

\section{Related Work}
\label{sec:related}
SymbolicLight V1 \citep{liu2026symboliclightv1} and its spike-aware CPU follow-up \citep{liu2026spikeaware} motivate the transition from trainable sparse activations to executable sparse inference. SpikeGPT \citep{zhu2023spikgpt} studies directly trained spiking language modeling, while SpikeLM \citep{xing2024spikelm} studies spike-driven language modeling with elastic bi-spiking. V2 continues V1's hybrid design and extends event computation while retaining continuous-state processing. Its focus here is the physical execution and energy of a trained checkpoint on an ARM edge CPU and a sparse FPGA.

Hybrid neuromorphic hardware provides a complementary architectural context. Tianjic \citep{pei2019tianjic} supports artificial and spiking neural-network computation on a fully digital platform with reconfigurable building blocks and hybrid coding. V2 uses the hybrid principle within a language model: sparse events coexist with recurrent state and a continuous residual, and its U50C FPGA prototype maps their computation and data movement into end-to-end inference. The measured energy benefit follows from the implemented execution mechanisms, rather than from biological analogy.

Integer-only inference \citep{jacob2018quantization,kim2021ibert} provides the broader setting for deterministic low-precision execution. Our integer reference supplies a common numerical target for successive FPGA implementations, with quantization quality evaluated separately. ALiBi \citep{press2022train} provides the position-bias mechanism used by the event attention path.

FPGA text-generation systems such as DFX \citep{hong2022dfx} and FlightLLM \citep{zeng2024flightllm} investigate customized hardware execution and mapping for language models. V2 studies how event-coded projection inputs enable active-row gathering, with measurements of both application throughput and card energy. Its architecture and workload differ from these systems, precluding direct cross-paper efficiency ranking. The supplementary GPU implementation uses the compilation system of PyTorch 2 \citep{ansel2024pytorch2}.

\section{Conclusion}
SymbolicLight V2 advances V1's spike-gated dual-path design into a hybrid neuromorphic language architecture with executable sparse inference. Sparse events and continuous-state processing work together in an end-to-end U50C FPGA prototype and an ARM integer implementation. Holding model weights fixed, selective weight access and valid-state KV loading increase short-context FPGA decode throughput by 35.5\% while reducing estimated gross card decode energy by 27.6\%. Complete-request energy falls by 24.4--27.7\% across three prefixes. An independent idle split shows how reduced token latency lowers the platform energy allocated to each output. AC-input measurements on four ARM cores additionally demonstrate local inference at approximately 9--10 W. For the same integer model, the U50C deployment also generates faster than the ROCK~5T CPU implementation, with lower gross and idle-subtracted incremental energy per token within the tested workloads and respective measurement boundaries.

These mechanisms provide a basis for V2 implementations on other dedicated hybrid neuromorphic hardware. Nonzero-event computation and memory access, valid-state transfers, and on-device generation are transferable; when their throughput gain exceeds the relative change in active power, other compatible implementations can likewise achieve faster generation and lower energy per token. U50C validates this design path, while the realized benefit depends on compute resources, memory organization, and control overhead.

\bibliographystyle{plainnat}
\bibliography{references_v2}

\appendix
\small
\section{Model Quality}
\label{app:quality}
\subsection{Same-budget architecture comparison}
Three 194M-scale models were trained from scratch using the same open training data, tokenizer with a 48K vocabulary, random seed, batch settings, and 4B-token budget. The dense GPT-2 control has 22 layers; the two models based on V2 have 12 layers. The continuous-attention control, denoted \czero{}, disables the event encoders added in V2 and uses continuous Q/K/V with softmax attention, while retaining the binary-spike sites inherited from V1. It is therefore a control for the added V2 mechanisms, not an entirely non-spiking model.

\begin{table}[!htbp]
\centering\small
\caption{FP32 quality of the trained models. PPL denotes perplexity and BPB bits per byte. Programmatic and downstream accuracies use byte-normalized continuation scoring. The hardware implementations use the same V2 checkpoint.}
\label{tab:quality}
\begin{tabular}{@{}lrrrr@{}}
\toprule
\bfseries Model & \bfseries PPL & \bfseries Natural BPB & \bfseries Programmatic acc. & \bfseries Downstream macro acc.\\
\midrule
\gnew{} & 12.502221 & 1.736573 & 0.335938 & 0.368302\\
\czero{} & 12.851092 & 1.774259 & 0.434896 & 0.362301\\
\snew{} & 14.749522 & 1.820616 & 0.234375 & 0.356452\\
\bottomrule
\end{tabular}
\end{table}
The programmatic suite contains 768 four-way items generated after training, covering arithmetic, symbolic ordering, progressions, and Python integer expressions. The natural-text test set contains 90 texts whose sources are separate from the training data; the downstream aggregate covers PIQA \citep{bisk2020piqa}, HellaSwag \citep{zellers2019hellaswag}, and ARC-Easy \citep{clark2018arc}. Relative to the same-topology control, V2's programmatic accuracy drops by 20.05 percentage points, failing the prespecified key-task criterion: each key task must retain at least 80\% of the continuous control's score and lose no more than 10 percentage points. Its full-model PPL increase over GPT-2 also exceeds the prespecified 15\% threshold for stopping further model scaling. These results fail the prespecified quality criteria, and the hardware changes do not alter that outcome. Programmatic accuracy of 0.234375 is below the four-way chance level of 0.25; the downstream macro accuracies of all three models are close to the aggregate chance level of approximately 0.33, limiting their separation at this scale.

The separate \srepair{} research variant restores causal softmax and removes the second projected Q/K/V eventization. Its PPL on the main held-out set is 13.426754, 4.48\% above \czero{}, but it is not the checkpoint deployed in these FPGA measurements. Its result is not combined with V2 energy to construct an equal-quality efficiency ratio.

\subsection{Quantization diagnostic and its scope}
A separate 5,657-token test set compares FP32 with the software integer reference. PPL is 90.19 versus 91.65, BPB 1.9787 versus 1.9857, top-1 accuracy 30.32\% versus 29.84\%, and top-5 accuracy 47.80\% versus 47.73\%. The relative PPL increase is approximately 1.6\%; greedy argmax agreement is 72.65\%. This panel is distinct from the main held-out set, so its absolute PPL is not directly comparable with \Cref{tab:quality}.

These are software-reference diagnostics. The software reference uses the training attention-window configuration, whereas the deployed FPGA build uses the extended deployment ring and its bitstream does not expose full logits. The panel therefore does not measure deployed perplexity across deployment context lengths. Token and state agreement with the integer deployment reference supplies a separate correctness check.

\section{Supplementary GPU Measurements}
\label{app:gpu}
\subsection{GPU protocol and precision boundary}
The RTX~5090 baseline uses compiled FP32, PyTorch 2.8.0/CUDA 12.8, TF32 disabled, KV caching, batch one, and three rounds. FPGA measurements followed five days later with the same weights, inputs, output lengths, and greedy decoding, but integer arithmetic. The FPGA window is 475 (480 slots including anchors), versus 256 on GPU. p32/n128 fits both windows; p256/n128 crosses the training window, giving different visible contexts. Identical FP32 and integer output sequences are not required.

Additional tests covered eager FP32, compiled BF16, TorchAO weight-only INT8/INT4, dynamic FP8, and custom cuBLASLt INT8. Custom INT8 succeeded. TorchAO dynamic W8A8 failed because its selected kernel required more than 16 matrix rows, while single-stream decode supplied one; this does not imply that the GPU lacks INT8 support. None exceeded compiled FP32 throughput in batch-one screening, although BF16 and weight-only INT8 used less energy in the separate experiment of \Cref{tab:gpu_precision}. CUDA graph capture attempts failed; TensorRT-LLM was not tested. The 9.15$\times$ energy ratio compares against the fastest tested FP32 baseline, not the lowest-energy GPU implementation.

Gross card energy is average active sensor power divided by generated-token throughput. The focused FPGA rerun uses XRT/\texttt{xbutil} telemetry at approximately 0.5\,Hz; the GPU measurement uses NVML at 10\,Hz. Runs last at least 60 seconds. Host and AC-supply costs are excluded; the sensors have not been cross-calibrated with an external meter.

\subsection{Additional decode results}
The principal comparison is in \Cref{sec:gpu_energy,tab:main}. At p32, FPGA throughput has a three-round spread of 0.04\%; an independent measurement gives 643.5 tokens/s, close to the focused rerun of 643.2. No matched p480 GPU row was measured.

\subsection{Prompt ingestion}

\begin{table}[!htbp]
\centering\small
\caption{Whole-request generated-token throughput and power-based gross energy estimates. The denominator includes prefill; the numerator remains 128 generated tokens. U50C power is taken from the corresponding continuous run, rather than separately integrated over each request.}
\label{tab:request}
\begin{tabular}{@{}lrrrrr@{}}
\toprule
 & \multicolumn{2}{c}{\bfseries U50C \implkv{}} & \multicolumn{2}{c}{\bfseries RTX 5090 FP32} & \bfseries Energy ratio\\
\bfseries Prefix & \bfseries \tps{} & \bfseries J/token & \bfseries \tps{} & \bfseries J/token & \bfseries GPU/FPGA\\
\midrule
32 & 518.4 & 0.05468 & 406.5 & 0.40466 & 7.40$\times$\\
128 & 304.8 & 0.09274 & 408.7 & 0.41878 & 4.52$\times$\\
256 & 187.3 & 0.14989 & 404.4 & 0.42522 & 2.84$\times$\\
\bottomrule
\end{tabular}
\end{table}
At p256, sequential FPGA prompt ingestion takes 424.6\,ms; the GPU's batched prompt forward has a recorded time to first token of 3.25\,ms. These intervals differ, so their quotient is not a speedup. \Cref{tab:request} compares complete requests including prefill.

\begin{samepage}
\subsection{Paired GPU architecture control}

A separate paired experiment on another RTX~5090 compares the V2 topology with its added event and attention changes disabled or enabled. With compiled FP32, \czero{} reaches 353.9~\tps{} and 0.4706\,J/token, versus 319.8~\tps{} and 0.5065\,J/token for \snew{}: throughput falls 9.6\% and energy rises 7.6\%. The normalizer and projected-event path change together, so this is not an isolated event-encoder ablation. Added zeros do not automatically improve the tested dense GPU execution.

The precision comparison from this same complete-request campaign is reported in \Cref{tab:gpu_precision}. BF16 and weight-only INT8 reduce card power and energy despite lower throughput. The table specifies the cross-prefix aggregation and the use of another GPU, preventing these results from being conflated with the baseline in \Cref{tab:main}.

\end{samepage}
\end{document}